\documentclass{article}
\usepackage{spconf,amsmath,graphicx}

\usepackage{tabularx}
\usepackage{arydshln}
\usepackage{cite}
\usepackage{textcomp}
\usepackage{placeins}
\usepackage{multirow}
\usepackage{amsmath,amssymb,amsfonts}
\usepackage{algorithmic}
\usepackage{algorithm}
\usepackage{graphicx}
\usepackage{subcaption}
\usepackage{textcomp}
\usepackage{placeins}
\usepackage{lipsum}
\usepackage{mathtools}
\usepackage{cuted}
\usepackage{xcolor}
\usepackage{times}
\usepackage{epsfig}
\usepackage{graphicx}
\usepackage{epstopdf}
\usepackage{enumitem}
\usepackage{fixltx2e}
\usepackage{array}
\usepackage[colorlinks=true]{hyperref}

\usepackage{times}
\usepackage{epsfig}
\usepackage{graphicx}
\usepackage{amsmath}
\usepackage{amssymb}
\usepackage{tabularx}
\usepackage{algorithmic}
\usepackage{algorithm}
\usepackage{textcomp}
\usepackage{placeins}
\usepackage{epstopdf}
\usepackage{multirow}
\usepackage{siunitx}
\usepackage{makecell,multirow}
\usepackage[font=small,skip=0pt]{caption}

\usepackage{xcolor}
\usepackage[utf8]{inputenc}

\title{Distortion Robust Image Classification using Deep Convolutional Neural Network with Discrete Cosine Transform}
\name{Md Tahmid Hossain \quad Shyh Wei Teng \quad Dengsheng Zhang \quad Suryani Lim \quad Guojun Lu\thanks{Published as a conference paper at ICIP 2019}}
\address{School of Science, Engineering and Information Technology\\
	Federation University, Gippsland Campus, Churchill, VIC 3842, Australia}
\begin{document}
%
\maketitle
\begin{abstract}
 Convolutional Neural Networks are highly effective for image classification. However, it is still vulnerable to image distortion. Even a small amount of noise or blur can severely hamper the performance of these CNNs. Most work in the literature strives to mitigate this problem simply by fine-tuning a pre-trained CNN on mutually exclusive or a union set of distorted training data. This iterative fine-tuning process with all known types of distortion is exhaustive and the network struggles to handle unseen distortions. In this work, we propose distortion robust DCT-Net, a Discrete Cosine Transform based module integrated into a deep network which is built on top of VGG16 \cite{vgg1}. Unlike other works in the literature, DCT-Net is `blind' to the distortion type and level in an image both during training and testing. The DCT-Net is trained only once and applied in a more generic situation without further retraining. We also extend the idea of dropout and present a training adaptive version of the same. We evaluate our proposed DCT-Net on a number of benchmark datasets. Our experimental results show that once trained, DCT-Net not only generalizes well to a variety of unseen distortions but also outperforms other comparable networks in the literature.
\end{abstract}

\section{Introduction}

\label{sec:intro}

\begin{figure*}[t!]
\centering
   \includegraphics[width=\textwidth]{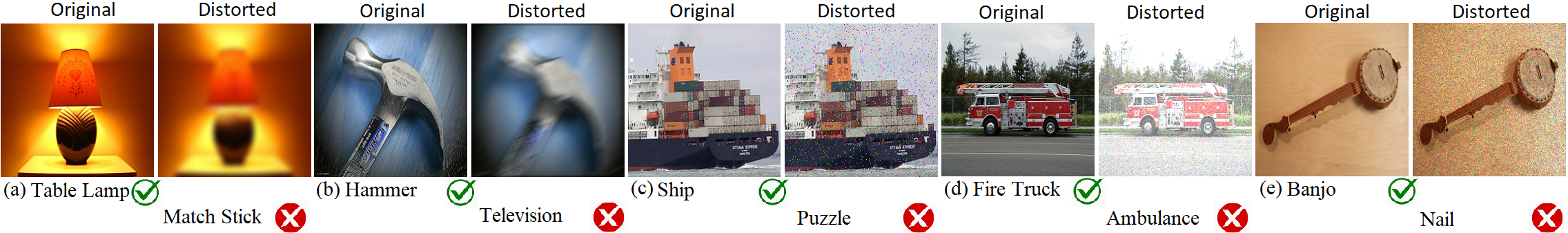}
   \caption{In a set of original and distorted sample pairs, VGG16 missclassifies all where the distortion types are (a) Gaussian blur (b) Motion blur (c) Salt and Pepper noise (d) Gaussian noise (e) Speckle noise. Proposed DCT-Net classifies all the distorted images correctly. }
\label{fig:comparison1}

\end{figure*}

The ImageNet challenge \cite{russakovsky1} has induced a number of new deep CNN architectures which are extremely effective at different computer vision tasks \cite{alexnet1,zf1,vgg1,resnet1,google1}. However, these deep networks are found to be susceptible to image distortion when classifying images \cite{icccn1,akaram1,anguyen1,RN95,RN101,RN102}. Figure \ref{fig:comparison1} provides visual evidence of this claim as VGG16 miss-classifies a set of distorted input. Most work in the literature tackle this problem using data augmentation or fine-tuning of networks by training models with images of each chosen distortion in a mutually exclusive manner. While this increases the accuracy of the network, its performance is still inferior to models trained on a single distortion type \cite{akaram1}. Moreover, the fact that the network has to be fine-tuned on all possible distorted image data separately makes it even more undesirable. All these facts culminate to an intriguing question.\\\textbf{\textit{Is it possible to attain such a robust network which only needs to be trained once and can then be applied to different datasets without further retraining?}}

\begin{figure}[t!]

\begin{center}
   \includegraphics[width=.7\linewidth]{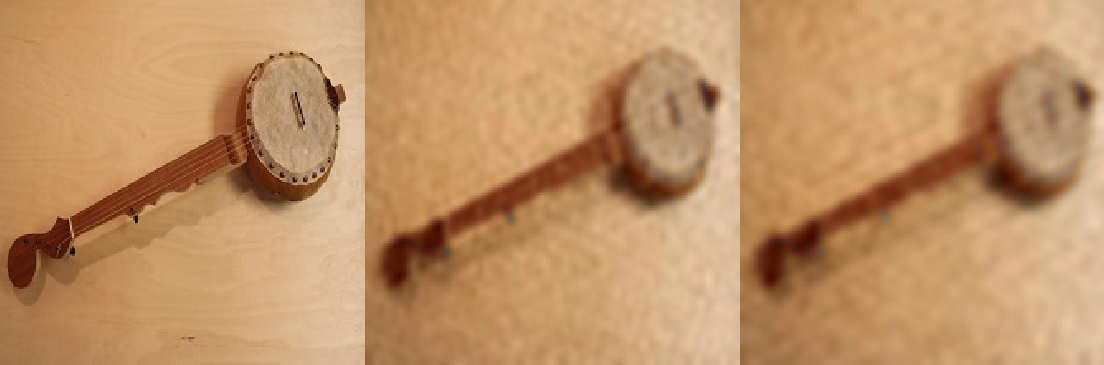}
\end{center}
   \caption{Sample output of DCT module (a-b). Each input is transformed into frequency domain via forward DCT and information is discarded in increasing magnitude from left to right.}
\label{fig:tiledct}
\end{figure}

 In this work, we propose Discrete Cosine Transform based deep network DCT-Net. We show that after the input is transformed into frequency domain, dropping DCT coefficients or certain frequency components helps the deep network to learn a more robust representation of the training images, which leads to a quality invariant CNN. The input diversity stemming from the integrated DCT module provides an assorted visual representation of the training data and the network gets to learn features from a wide array of variants of a single image. DCT-Net no longer heavily relies on minute image details for learning and therefore when faced with a degraded version of an image, it can still classify correctly.
DCT module eliminates certain DCT coefficients based on a randomly chosen threshold per image. The threshold value \textit{X} is selected for each training input from a standard uniform distribution of integers (\textit{X $\sim U[0,50]$}). When \textit{X} happens to be 0, the image does not get transformed at all and is fed forward in the input's original form. Values close to 50, on the other hand, remove more details from the image. Since most of the discarded DCT coefficients are of high frequencies, discarding them contributes to loss of image details as depicted in Figure \ref{fig:tiledct}. Since the loss of input information is random for each image, every epoch of our training yields a different representation of each input.

We evaluate our network on Gaussian noise and blur, salt and pepper noise, motion blur and speckle.
This paper is organized as follows: Section \ref{motivation} explains our motivation. Section \ref{related} discusses related works. Section \ref{dct} introduces and formulates the proposed DCT-Net with adaptive dropout. Section \ref{results} describes the benchmark datasets and provides a detailed performance comparison. Section \ref{conclusion} concludes this paper.

\section{Motivation} \label{motivation}

 Most of the well-established deep networks presume that the input images are free of artifacts. However, in real-life scenarios, images can be distorted in various ways e.g. images captured in low light can be noisy, Motion/Gaussian blur can occur if the camera or subject is moving. In transmission, packet-loss can potentially result in missing image details, noise, or missing frequencies. Sometimes, surveillance images are taken under challenging weather conditions or the device used is of substandard quality. All these factors can result in degraded visual data. In addition to classification, distortion can also degrade the performance of a CNN based system in other computer vision tasks e.g. object detection, image registration, segmentation and even more complex tasks like autonomous car driving, facial recognition etc. The aforementioned factors inspired us to explore ways of overcoming the drawbacks of traditional CNN. We aim to solve this problem by getting DCT-Net to learn a more robust representation of the training data.

\section{Related Work}\label{related}

Despite the deep networks' vulnerability to distortion or quality degradation, few approaches can be found in the literature dealing with this problem. A simple approach to add robustness to neural networks is to fine-tune the network on images with the expected distortions. Vasiljevic \textit{et al.} \cite{vas1} show that this approach works well for blurred images. They achieve satisfactory performance on clean and blurred testsets. Similarly, Zhou \textit{et al.} \cite{zhou1} show the effectiveness of fine-tuning for both noisy and blurry images. Interestingly, the model trained on both noisy and blurry images has a much higher error rate than the average error rate of models trained only on noise and blur when these latter models are tested on their respective distorted or degraded testsets. Our Results section will shed further light on this aspect. Dodge \textit{et al.} \cite{akaram1} propose a mixture of experts-based model for image classification (MixQualNet). Their network consists of expert networks where the experts are trained for particular distortion types and a gating network which is trained to select among the experts. This model shows better performance compared to \cite{zhou1,vas1}. However, MixQualNet is basically a complex ensemble of \textit{N} number of identical CNN models where \textit{N} is the number of distortion types the model is trained on. This already parameter heavy model has one million additional gating network parameters, all resulting in a sluggish training process. Diamond \textit{et al.} \cite{diamond1} and Yim \textit{et al.} \cite{RN95} propose a system which modifies a neural network with an additional layer for undistorting the images by denoising and deblurring. For these reconstructed images, further fine-tuning of a deep neural network is performed. The method has prior knowledge of camera noise and blurring parameters, but for general applications (e.g. images from the internet) the camera parameters may be unavailable. This drawback greatly limits the applicability of the method.

To reduce data overfitting during training, constant dropout probability \cite{dropout1,Baheti1} is used in the literature. The dropout probability (\textit{P}) used in VGG16 \cite{vgg1} is 0.5. Our results in Section \ref{results} will show that the proposed training adaptive incremental dropout enables the deep network to generalize better. 

\section{DCT-Net}\label{dct}
\subsection{Discrete Cosine Transform}
DCT is a popular technique to analyze signals in the frequency domain. It is widely used in numerous applications, from lossy compression of audio (e.g. MP3) and image (e.g. JPEG) to spectral methods for the numerical solution of partial differential equations. To perform the Forward DCT (FDCT) in a standard JPEG compression \cite{wallace1}, each image is divided into $8\times8$ blocks, i.e. a 64-point discrete signal. However, it is found that this block-wise DCT operation may lead to undesired properties like blocking artifacts \cite{wang2000blind,chou1998simple}. Therefore, we consider only one block with dimensions equivalent to the height (\textit{H}) and width (\textit{W}) of the original input image. FDCT takes $H \times W$ signal as its input and outputs the corresponding basis-signal amplitudes or “DCT coefficients”. The DCT coefficient values can thereby be regarded as the relative amount of the 2D spatial frequencies contained in the original input signal, which in our case is an image. 
One of the most important features of FDCT is that it concentrates most of the signal energy in a few transformed DCT coefficients in the lower spatial frequencies \cite{wallace1}. In other words, the number of DCT coefficients with substantially high magnitude is very low and the smaller coefficients are far greater in number. More often than not, majority of the information in a natural image is represented in low frequencies. High frequencies generally encode sharp changes that add extremely fine details to the image.

There are a number of ways to perform DCT in the literature. We make use of Fast Cosine Transform \cite{makhoul1,narasimha1} because of its computational efficiency (\textit{NLogN}). We use Equation \ref{eq:1} on an input image A for FDCT and Equation \ref{eq:2} for Inverse DCT (IDCT) to obtain the reconstructed image B.

\noindent
\begin{equation}\label{eq:1}
\noindent
\small
\noindent
\begin{aligned}
B_{pq}=\alpha_{p}\alpha_{q}\sum\limits_{m=0}^{M-1}\sum\limits_{n=0}^{N-1}A_{mn}
\cos{\frac{\pi(2m +1)p}{2M}}
\cos{\frac{\pi(2n +1)q}{2N}},\\
0\leq p \leq M-1, 0\leq q \leq N-1 
\end{aligned}
\end{equation}

\noindent
\begin{equation}\label{eq:2}
\small
\noindent
\begin{aligned}
A_{mn}=\sum\limits_{p=0}^{M-1}\sum\limits_{q=0}^{N-1}\alpha_{p}\alpha_{q}B_{pq}
\cos{\frac{\pi(2m +1)p}{2M}}
\cos{\frac{\pi(2n +1)q}{2N}},\\
0\leq m \leq M-1, 0\leq n \leq N-1 
\end{aligned}
\end{equation}
\noindent
\begin{equation*}
\small
  \alpha_p=\begin{cases}
    \sqrt{\frac{1}{M}}, & p = 0.\\
    \sqrt{\frac{2}{M}}, & 1\leq p \leq M-1 
  \end{cases}
\end{equation*} 
\begin{equation*}
     \alpha_q=\begin{cases}
    \sqrt{\frac{1}{N}}, & q = 0.\\
    \sqrt{\frac{2}{N}}, & 1\leq q \leq N-1 
  \end{cases}
\end{equation*}
Here M and N are the row and column sizes of the input and output images respectively.


\begin{table*}[t!]
\small
\noindent
\caption{Performance comparison of different networks. Accuracy over each distorted testset is the mean over the specified distortion levels. Overall accuracy is the average of clean and distorted datasets. The best accuracy is displayed in bold and the second best is underlined.}
\label{table:1}

\begin{tabularx}{.98\textwidth}{p{3cm}|p{1cm}|p{1.7cm}|p{2cm}|p{1.3cm}|p{1.6cm}|p{1.6cm}|p{1.8cm}}
\hline
\multicolumn{8}{|c|}{CIFAR-10}\\
\hline
CNN Model & Original & Gauss. Noise & Salt and Pepper & Speckle & Gauss. Blur & Motion Blur & Overall Accu.\\
\hline
M\textsubscript{clean} (VGG16 \cite{vgg1}) & \textbf{88.43} & 40.53 & 42.60 & 45.10& 61.43 & 56.35 & 55.74 \\
\hline 
M\textsubscript{Gnoise} & 65.50 & \textbf{62.91} & 59.10 & 54.32 & 23.50 & 20.36 & 47.60\\
\hline
M\textsubscript{Gblur}\cite{vas1} & 68.84 & 29.33 & 30.25 & 38.36 & 73.36 & 50.51 & 48.44\\

\hline
M\textsubscript{BN} \cite{zhou1} & 83.36 & 59.06 & 51.87 & 55.08 & 69.36 & 52.98 & 61.95\\
\hline
M\textsubscript{All} & 80.54 & 57.95 & \underline{60.77} & \textbf{63.12} & 68.14 & 59.20 & 64.95\\
\hline
MixQualNet\cite{akaram1} & 81.56 & \underline{60.69} & 58.99 & 57.36 & 70.26 & 62.89 & 65.29\\
\hline
DCT-Net\textsubscript{NoAdaptiveDropout} & 85.36 & 55.87 & 59.02 & 61.39 & \underline{76.41} & \underline{68.51} & \underline{67.76}\\
\hline
DCT-Net & \underline{86.97} & 57.06 & \textbf{61.23} & \underline{62.05} & \textbf{77.15} & \textbf{70.91} & \textbf{69.23}\\
\hline


\multicolumn{8}{|c|}{CIFAR-100}\\
\hline
CNN Model & Original & Gauss. Noise & Salt and Pepper & Speckle & Gauss. Blur & Motion Blur & Overall Accu.\\
\hline
M\textsubscript{clean} (VGG16 \cite{vgg1}) & \textbf{67.49} & 25.52 & 29.62 & 32.91 & 50.39 & 48.30 & 42.37 \\
\hline 
M\textsubscript{Gnoise} & 52.54 & \textbf{49.80} & 59.22 & 41.20 & 17.61 & 14.25 & 39.10\\
\hline
M\textsubscript{Gblur}\cite{vas1} & 56.81 & 19.69 & 22.26 & 27.35 & \underline{62.37} & 42.77 & 38.54\\
\hline
M\textsubscript{BN} \cite{zhou1} & 62.44 & \underline{46.08} & 38.28 & 33.03 & 52.46 & 35.88 & 44.70\\
\hline
M\textsubscript{All} & 60.68 & 45.13 & 60.20 & \textbf{61.39} & 53.32 & 49.10 & 54.97\\
\hline
MixQualNet\cite{akaram1} & 63.50 & 45.89 & 58.01 & 57.61 & 55.22 & 49.83 & 55.01\\
\hline
DCT-Net\textsubscript{NoAdaptiveDropout} & 64.30 & 44.11 & \underline{60.25} & 60.39 & 62.32 & \underline{53.50} & \underline{57.48}\\
\hline
DCT-Net & \underline{65.39} & 44.85 & \textbf{62.33} & \underline{60.65} & \textbf{64.05} & \textbf{55.91} &  \textbf{58.86}\\
\hline


\multicolumn{8}{|c|}{ImageNet}\\
\hline
CNN Model & Original & Gauss. Noise & Salt and Pepper & Speckle & Gauss. Blur & Motion Blur & Overall Accu.\\
\hline
M\textsubscript{clean} (VGG16 \cite{vgg1}) & \textbf{92.60} & 32.45 & 15.32 & 24.98 & 29.53 & 25.67 & 36.76 \\
\hline 
M\textsubscript{Gnoise} & 49.60 & \textbf{55.31} & 40.21 & 33.55 & 11.95 & 10.66 & 33.55\\
\hline
M\textsubscript{Gblur}\cite{vas1} & 55.36 & 16.24 & 18.91 & 22.63 & 39.88 & 30.69 & 30.62\\
\hline
M\textsubscript{BN} \cite{zhou1} & 75.77 & \underline{54.25} & 42.30 & 38.65 & 30.14 & 27.85 & 44.83\\
\hline
M\textsubscript{All} & 74.12 & 49.22 & 45.35 & 44.22 & 38.32 & \underline{40.50} & 48.79\\
\hline
MixQualNet\cite{akaram1} & 72.85 & 50.68 & {49.63} & 47.34 & 39.33 & 34.98 & 49.14\\
\hline
DCT-Net\textsubscript{NoAdaptiveDropout} & 85.35 & 52.12 & \underline{50.66} & \underline{50.30} & \underline{42.25} & 39.77 & \underline{53.41}\\
\hline
DCT-Net & \underline{87.50} & 52.98 & \textbf{51.30} & \textbf{50.98} & \textbf{43.90} & \textbf{41.32} & \textbf{54.66}\\
\hline

\end{tabularx}


\end{table*}

\subsection{Network Architecture and Training}\label{module}

DCT module in the DCT-Net transforms each of the training images using FDCT and produces a set of DCT coefficients. A uniform random variable \textit{(X)} is chosen for each training image from a predefined range of 0 to 50 (\textit{X $\sim U[0,50]$}) which is effectively a random DCT coefficient threshold value. All DCT coefficients lying under the chosen threshold \textit{X} are turned to 0. When the random threshold is equal or close to 0, the input image undergoes very little or no transformation at all which means there is hardly any loss of image detail. On the other hand, if the random threshold happens to be a large number which is close to 50, this thresholding step removes most of the high frequencies from an image and leaves most of the low frequencies intact along with the DC component. Since a large part of the signal strength is stored in the low spectrum, the loss of information takes away mostly sharp changes and finer details of the image pertaining to different edges and contours. Along with the omission of most of the high frequencies, some of the low frequencies with little visual impact on the input image may get discarded as well in the process. The thresholding considers the absolute values of the coefficients. Inverse DCT or IDCT is performed on the remaining DCT coefficients to reconstruct the transformed image. This DCT transformed image is then fed forward to the convolutional layer. The DCT module is free of trainable parameters and no backpropagation based learning takes place within this module. Once the deep network is trained, this module is removed from the network and test images directly enter into the first convolutional layer.
 
 We employ VGG16 \cite{vgg1} as the base of our proposed DCT-Net. VGG16 uses $3\times3$ filters in all 13 convolutional layers, $2\times2$ max pooling with stride 2 and ReLU is used as activation function. All weights are initialized on ImageNet for our training. We use stochastic gradient descent with momentum 0.9, learning rate 0.01 and minibatch 128 (40 epochs).
 
 Dropout is a well-known technique to counter the effect of overfitting \cite{dropout1}. We extend the idea of dropout and deploy an adaptive version of it during training. The dropout probability \textit{P} is initialized with 0.1 and it is updated from a range of [0.1 - 0.5] with smallest increment unit of 0.1. The update only comes into effect when the network seems to converge to training data and the possibility of overfitting is inevitable. \textit{P} is updated when minibatch training accuracy reaches and stays above 80\% for an entire epoch. The value of \textit{P} depends on the remaining number of epochs which is divided into five equal intervals. As the training proceeds forward, \textit{P} increases in each of the five epoch intervals by 0.1.



\section{Datasets and performance Evaluation}\label{results}
We evaluate the networks on the clean data and five different types of distortion namely Gaussian noise, Gaussian blur, salt and pepper noise, motion blur and speckle noise. We consider CIFAR-10/100 \cite{cifar1} and ImageNet \cite{imagenet1} as our benchmark datasets. CIFAR-10 and CIFAR-100 consist of 60,000 images in 10 and 100 classes respectively (500,00 training, 100,00 testing). ImageNet has 1000 classes with 120,0000 training and 100,000 test images.
Gaussian noise is added with the variance range [0.1-0.5]. For Gaussian blur, the standard deviation range is [0-5] for CIFAR-10/100 and [0-10] for ImageNet. Salt and pepper noise is added to an image pixel with a probability range [0-0.5]. For motion blur in CIFAR-10/100, we use motion angle range between 0 to 22.5 degrees. For ImageNet, the motion angle range is 0 to 45 degrees. Variance range [0.1-0.5] is used for speckle noise. All the test images are tested at five different distortion levels uniformly chosen from the defined range of distortions.

We refer our model as DCT-Net in this work. We denote M\textsubscript{clean} as the VGG16 network trained on the original clean dataset. M\textsubscript{Gnoise} and M\textsubscript{Gblur} \cite{vas1} are effectively M\textsubscript{clean} fine-tuned on Gaussian noise and blur containing training set respectively. M\textsubscript{BN} \cite{zhou1} is fine-tuned on both Gaussian blur and noise. M\textsubscript{All} is the VGG16 network fine-tuned on all five distortions. MixQualNet is the ensemble of individual distortion expert models with a gating network \cite{akaram1}.
Table \ref{table:1} compares the classification accuracy of these base models. All of the deep networks are tested against five increasing levels of corresponding type of noise and blur. The mean accuracy over all distortion levels is displayed in each of the results' column of Table \ref{table:1} except the first and last one. For a particular network, the original accuracy in Table \ref{table:1} is computed over the corresponding clean testset whereas the overall accuracy is the numerical average over all six individual accuracies. It is worth noticing that the models fine-tuned on one specific type of distortion e.g.  M\textsubscript{Gnoise} and M\textsubscript{Gblur} \cite{vas1} perform well on their specific distorted testset. However, they fail to generalize well to other types of distortion and also accuracy on the clean test data drops. 
On the other hand,  M\textsubscript{BN} \cite{zhou1} has a mediocre performance on all the testsets and could not surpass  M\textsubscript{Gnoise} and M\textsubscript{Gblur} in their particular distortion testsets in any of the benchmark datasets. Moreover, the performance of M\textsubscript{BN} propels the belief that the poor performance of these networks on unseen distortion remains a major drawback.
M\textsubscript{All} and MixQualNet \cite{akaram1} displays competitive results in a number of testsets but the rigid training image requirement with all possible distortions should be kept in mind. The overall accuracy of M\textsubscript{All} and MixQualNet is 64.95\%, 54.97\%, 48.79\% and 65.29\%, 55.01\%, 49.14\% in CIFAR-10, CIFAR-100 and ImageNet respectively. This is consistent with the ways these networks are designed and trained.
Our proposed DCT-Net ouperforms others and copes well with all types of image distortion in all three datasets (69.23\% in CIFAR-10, 58.86\% in CIFAR-100 and 54.66\% in ImageNet). It also performs consistently on distorted dataset while negligible drop in performance on the original clean testset.

\section{Conclusion}\label{conclusion}
We have proposed DCT-Net with an adaptive dropout and have shown that discarding a part of the input signals or image details based on DCT adds diversity to each of the training data. Since the threshold used for discarding information is random for each image, every epoch is likely to produce a different version of a training image. Unlike traditional exhaustive fine-tuning, our network gets to learn a strong feature representation from all the variants stemming from each image. 
Although we focused mainly on classification, substandard image quality can adversely affect the performance of any other CNN-based computer vision task. Our proposed method can easily be applied to other existing networks which further adds to the contribution of this work.



{\small
\bibliographystyle{IEEEbib}
\bibliography{cvprRef}
}
\end{document}